\documentclass[a4paper,twoside]{article}

\usepackage{epsfig}
\usepackage{subcaption}
\usepackage{calc}
\usepackage{amssymb}
\usepackage{amstext}
\usepackage{amsmath}
\usepackage{amsthm}
\usepackage{multicol}
\usepackage{pslatex}
\usepackage{apalike}
\usepackage{pgfplots}
\usepackage{filecontents}

\usepackage{SCITEPRESS}     

\begin{document}

\title{Combating Mode Collapse in GAN training: An Empirical Analysis using Hessian Eigenvalues}

\author{\authorname{Ricard Durall\sup{1,2,4}, Avraam Chatzimichailidis\sup{1,3,4}, Peter Labus \sup{1,4} and Janis Keuper\sup{1,5}}
\affiliation{\sup{1}Fraunhofer ITWM, Germany}
\affiliation{\sup{2}IWR, University of Heidelberg, Germany}
\affiliation{\sup{3}Chair for Scientific Computing, TU Kaiserslautern, Germany}
\affiliation{\sup{4}Fraunhofer Center Machine Learning, Germany}
\affiliation{\sup{5}Institute for Machine Learning and Analytics, Offenburg University, Germany}}



\abstract{Generative adversarial networks (GANs) provide state-of-the-art results in image generation. However, despite being so powerful, they still remain very challenging to train. This is in particular caused by their highly non-convex optimization space leading to a number of instabilities. Among them, mode collapse stands out as one of the most daunting ones. This undesirable event occurs when the model can only fit a few modes of the data distribution, while ignoring the majority of them. In this work, we combat mode collapse using second-order gradient information. To do so, we analyse the loss surface through its Hessian eigenvalues, and show that mode collapse is related to the convergence towards sharp minima. In particular, we observe how the eigenvalues of the $G$ are directly correlated with the occurrence of mode collapse. Finally, motivated by these findings, we design a new optimization algorithm called nudged-Adam (NuGAN) that uses spectral information to overcome mode collapse, leading to empirically more stable convergence properties.}

\onecolumn \maketitle \normalsize \setcounter{footnote}{0} \vfill

\section{\uppercase{Introduction}}
\label{sec:introduction}

\noindent Although Deep Neural Networks (DNNs) have exhibited remarkable success in many applications, the optimization process of DNNs remains a challenging task. The main reason for that is the non-convexity of the loss landscape of such networks. While most of the research in the field has focused on single objective minimization, such as classification problems, there are other models that require the joint minimization of several objectives. Among these models, Generative Adversarial Networks (GANs) \cite{goodfellow2014generative} are particularly interesting, due to their success of learning entire probability distributions. Since their first appearance, they have been used to improve the performance
of a wide range of tasks in computer vision,
including image-to-image translation \cite{abdal2019image2stylegan,karras2020analyzing},
image inpainting \cite{iizuka2017globally,yu2019free},
semantic segmentation \cite{xue2018segan,durall2019object} and many more.

\let\thefootnote\relax\footnotetext{Accepted in VISAPP 2021}

GANs are a class of generative models which consist of a generator ($G$) and a discriminator ($D$) DNN model. Within an adversarial game they are trained in such a way that the $G$ learns to produce new samples distributed according to the desired data distribution.
Training can be formulated in terms of a minimax optimization of a value function $V(G, D)$
\begin{align}
\begin{split}
	\min_G \max_D \; V(G,D).
\end{split}
\end{align}
While being very powerful and expressive, GANs are known to be notoriously hard to train.
This is because their training is equivalent to the search of Nash equilibria in a high-dimensional, highly non-convex optimization space. The standard algorithm for solving this optimization problem is gradient descent-ascent (GDA), where $G$ and $D$ perform alternating update steps using first order gradient information w.r.t. the loss function. In practise, GDA is often combined with regularization, which has yield many state-of-the-art results for generative models on various benchmark datasets. However, GDA is known to suffer from undesirable convergence properties that may lead to instabilities, divergence, catastrophic forgetting and mode collapse. The latter term refers to the scenario where only a few modes of the data distribution are generated and the model produces only a limited variety of samples.

Many recent works have studied different approaches to tackle these issues. In reference \cite{radford2015unsupervised}, for instance, was one of the first attempts to use convolutional neural networks in order to improve both training stability, as well as the visual quality of the generated samples. Other works achieved improvements through the use of new objective functions \cite{salimans2016improved,arjovsky2017wasserstein}, and additional regularization terms \cite{gulrajani2017improved,durall2020watch}.
There have also been recent advances in the theoretical understanding of GAN training.
References \cite{nagarajan2017gradient,mescheder2017numerics}, for examples, have investigated the convergence properties of GAN training using first-order information.
There, it has been shown that a local analysis of the eigenvalues of the Jacobian of the loss function can provide guarantees on local stability properties.
Moreover, going beyond first order gradient information, references \cite{berard2019closer,fiez2019convergence} have used the top k-eigenvalues of the Hessian of the loss to investigate the convergence and dynamics of GANs.

In this paper, we conduct an empirical study to obtain new insights concerning stability issues of GANs.
In particular, we investigate the general problem of finding local Nash equilibria by examining the characteristics of the Hessian eigenvalue spectrum and the geometry of the loss surface.  We thereby verify some of the previous findings that were based on the top k-eigenvalues alone. We hypothesize that mode collapse might stand in close relationship with convergence towards sharp minima, and we show empirical results that support this claim.  Finally, we introduce a novel optimizer which uses second-order information to combat mode collapse. We believe that our findings can contribute to understand the origins of the instabilities encountered in training GAN.

In summary our contributions are as follows
\begin{itemize}
\item We calculate the full Hessian eigenvalue spectrum during GAN training, allowing us to link mode collapse to anomalies of the eigenvalue spectrum.
\item We identify similar patterns in the evolution of the eigenvalue spectrum of $G$ and $D$ by inspecting their top k-eigenvalues.
\item We introduce a novel optimizer that uses second-order information to mitigate mode collapse.
\item We empirically demonstrate that $D$ finds a local minimum, while $G$ remains in a saddle point.  
\end{itemize}

\section{\uppercase{Related Work}}

\noindent While gradient-based optimization has been very successful in Deep Learning, applying gradient-based algorithms in game theory, i.e. finding Nash equilibria, has often highlighted their limitations. An intense line of research based on first- and second-order methods has studied the dynamics of gradient descent-ascent by investigating the loss landscape of DNNs. One of the initial first-order approaches \cite{goodfellow2014qualitatively} studied the properties of the loss landscape along a linear path between two points in parameter space. In doing so, it was demonstrated that DNNs tend to behave similarly to convex loss functions along these paths.
In later references \cite{draxler2018essentially} non-linear paths between two points were investigated. There, it was shown that the loss surface of DNNs contains paths that connect different minima, having constant low loss along these paths.

In the context of second-order approaches, there has also been notable progress \cite{sagun2016eigenvalues,alain2019negative}. There, the Hessian w.r.t. the loss function was used to reduce oscillations around critical points in order to obtain faster convergence to Nash equilibria. The main advantage of second-order methods is the fact that the Hessian provides curvature information of the loss landscape in all directions of parameter space (and not only along the path of steepest descent as with first-order methods). However, this curvature information is local only and very expensive to compute. In the context of GANs, second-order methods have not been investigated in depth. Recent works \cite{berard2019closer,fiez2019convergence} have not calculated the full Hessian matrix but resorted to approximations, such as computing the top-k eigenvalues only. To the best of our knowledge, we are the first to use the full Hessian eigenvalue spectrum to study the training behavior of GANs.

Another line of research tries to classify different types of local critical points of the loss surface during training w.r.t. their generalization behavior. In \cite{hochreiter1997flat} it was originally speculated that the width of an optimum is critically related to its generalization properties. Later, \cite{keskar2016large} extended the conjectures by conducting a set of experiments showing that SGD usually converges to sharper local optima for larger batch sizes. Following this principle, \cite{chaudhari2019entropy} proposed an SGD-based method that explicitly forces optimization towards wide valleys. \cite{izmailov2018averaging} introduced a novel variant of SGD which averages weights during training. In this way, solutions in flatter regions of the loss landscape could be found which led to better generalization.
This, in turn, has led to measurable improvements in applications such as classification.
However, \cite{LaurentSharp} argues that the commonly used definitions of flatness are problematic. By exploiting symmetries of the model, they can change the amount of flatness of a minimum without changing its generalization behavior.

\section{\uppercase{Preliminaries}}

\subsection{Formulation of GANs}
\noindent The goal of a generative model is to approximate a real data distribution $p_{\mathrm{r}}$ with a surrogate data distribution $p_{\mathrm{f}}$.
One way to achieve this is to minimize the ``distance'' between those two distributions.
The generative model of a GAN, as originally introduced by
Goodfellow\textit{ et al.}, does this by minimizing the Jensen-Shannon Divergence between $p_{\mathrm{r}}$ and $p_{\mathrm{f}}$, using the feedback of the $D$.
From a game theoretical point of view, this optimization problem may be seen as a zero-sum game between two players, represented by the discriminator model and  the generator model, respectively.
During the training, the $D$ tries to maximize the probability of correctly classifying a given input as real or fake by updating
its loss function
\begin{align}
\begin{split}
	\mathcal{L}_D = \mathbb{E}_{\mathrm{\mathbf{x}} \sim p_{\mathrm{r}}} \left[ \log \left(D(\mathrm{\mathbf{x}})\right) \right] \,+ \, \mathbb{E}_{\mathrm{\mathbf{z}} \sim p_{\mathrm{z}}}[\log(1-D(G(\mathrm{\mathbf{z}}))], 
\end{split}
\end{align}
through stochastic gradient ascent.
Here, $\textbf{x}$ is a data sample and $\textbf{z}$ is drawn randomly.

The $G$, on the other hand, tries to minimize the probability of $D$ to classify its generated data correctly. This is done by updating its loss function
\begin{align}
\begin{split}
	\mathcal{L}_G = \mathbb{E}_{\mathrm{\mathbf{z}} \sim p_{\mathrm{z}}}[\log(1-D(G(\mathrm{\mathbf{z}}))]
\end{split}
\label{for:gen}
\end{align}
via stochastic gradient descent. 
As a result, the joint optimization can be viewed as a minimax game between $G$, which learns how to generate new samples distributed according to $p_{\mathrm{r}}$, and $D$, which learns to discriminate between real and generated data. The equilibrium of this game is reached when the $G$ is generating samples that look as if they were drawn from the training data, while the $D$ is left indecisive whether its input is generated or real.

\subsection{Training of GANs}

\noindent As we explained above, the training of a GAN requires the joint optimization of several objectives making their convergence intrinsically different from the case of a single objective function. The optimal joint solution to a minimax game is called Nash-equilibrium.
In practice, since the objectives are non-convex, using local gradient information, we can only expect to find local optima, that is local Nash-equilibria (LNE) \cite{adolphs2018local}.
An LNE is a point for which there exists a local neighborhood in parameter space, where neither the $G$ nor the $D$ can unilaterally decrease/increase their respective losses,
i.e. their gradients vanish while their second derivative matrix is positive/negative semi-definite:
\begin{equation}
  \begin{gathered}
    	||\nabla_{\theta} \mathcal{L}_G|| = ||\nabla_{\varphi} \mathcal{L}_D|| = 0,\\
	    \nabla_\theta^2 \mathcal{L}_G \succeq 0 \; \mathrm{and} \; \nabla_\varphi^2 \mathcal{L}_D \preceq 0
  \end{gathered}
  \label{eqn:LNE}
\end{equation}
Here, $\theta$ and $\varphi$ are the weights $D$ and $G$, respectively.

\begin{figure}[!t]
\centering
   \includegraphics[width=\linewidth]{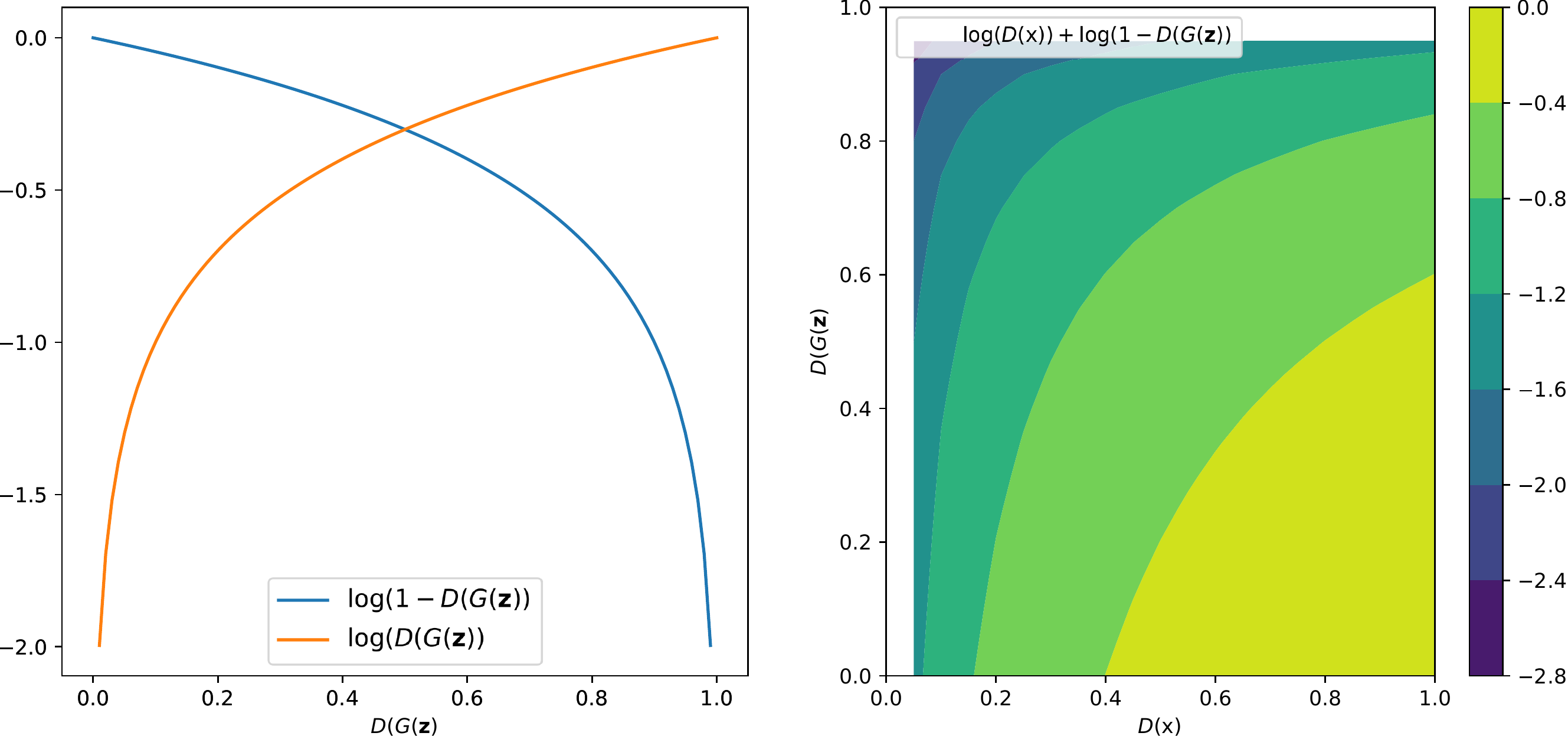}
      \caption{(Left) $G$ losses, either minimization of $\log(1-D(G(\mathrm{\mathbf{z}}))$ or maximization of 
      $\log(D(G(\mathrm{\mathbf{z}}))$. (Right) $D$ loss, maximization of $\log(D(\mathrm{x})) + \log(1-D(G(\mathrm{\mathbf{z}}))$.}
\label{fig:nsgan}
\end{figure}

\subsection{Evaluation of GANs}

GANs have experienced a dramatic improvement in terms of image quality in recent years. Nowadays, it is possible to generate artificial high resolution faces indistinguishable from real images to humans \cite{karras2020analyzing}. However, their evaluation and comparison remain a daunting task, and so far, there is no consensus as to which metric can best capture strengths and limitations of models. Nevertheless, the Inception Score (IS), proposed by \cite{salimans2016improved}, is the most widely adopted metric. It evaluates images generated from the model by determining a numerical value that reasonably correlates with the quality and diversity of output images. 

\section{\uppercase{Method}}
\subsection{Non-Saturating GAN}
\noindent In minimax GANs the $G$ attempts to generate samples that minimize the probability of being detected as fake by the $D$ (c.f. Formula~\ref{for:gen}). However, in practice, it turns out to be advantages to use an alternative cost function which instead ensures that the generated samples have a high probability of being considered real. This modified version of a GAN is called a non-saturating GAN (NSGAN). When training a NSGAN the $G$ maximizes an alternative objective
\begin{align}
\begin{split}
	\mathbb{E}_{\mathrm{\mathbf{z}} \sim p_{\mathrm{z}}}[\log(D(G(\mathrm{\mathbf{z}}))].
\end{split}
\end{align}
The intuition why NSGANs perform better than GANs is as follows.
In case the model distribution is highly different from the data distribution, the NSGAN can bring the two distributions closer together since the loss function generates a strong gradient. In fact, the NSGAN will have a vanishing gradient only when the $D$ starts being indecisive whether its input is from the data distribution or the $G$. This is acceptable, however, since the samples will already have reached the distribution of the real data by that time. 
Figure~\ref{fig:nsgan} shows the loss function of the original and non-saturating  $D$ and $G$, respectively.

\subsection{Stochastic Lanczos quadrature algorithm} 
\noindent During optimization the network tries to converge into a local minimum of the loss landscape. Here we differentiate between flat and sharp minima. Whether a minimum is considered sharp or flat is determined by the loss landscape around the converged point. If the region has approximately the same error, the minimum is considered flat, otherwise we refer to the minimum as sharp. Another method to determine the sharpness of a minimum is by looking at the eigenvalues of the Hessian. These describe the local curvature in every direction of the parameter space. This allows us to see whether our network converges into sharp or flat minima or whether it converges into a minimum at all. Here, big eigenvalues correspond to a sharp minimum in the corresponding eigendirection.

We observe that high eigenvalues in the $G$ and $D$ lead to a worse IS score. Therefore we conclude that mode collapse is linked to the network converging into sharp minima. In order to confirm this, we look at the full eigenvalue density spectrum during training. 

Calculating the eigenvalues of the Hessian has a complexity of $O(N^3)$, and storing the Hessian itself in order to compute the eigenvalues scales with $O(N^2)$ where $N$ is the number of parameters in the network. For neural networks that typically have millions of parameters, calculating the eigenvalues of their Hessian is infeasible. We can skip the problem of storing the Hessian by only calculating the Hessian-vector product for different vectors. In combination with the Lanczos algorithm, this allows us to compute the eigenvalues of the Hessian without having to calculate and store the Hessian itself.

The stochastic Lanczos quadrature algorithm \cite{lanczos1950iteration} is a method for the approximation of the eigenvalue density of very large matrices. The eigenvalue density spectrum is given by
\begin{equation}
\phi(t) = \frac{1}{N}\sum_{i=1}^{N}\delta(t-\lambda_i)
\end{equation}
where $N$ is the number of parameters in the network, $\lambda_i$ is the i-th eigenvalue of the Hessian and $\delta$ is the Dirac delta function.
In order to deal with the Dirac delta function the eigenvalue density spectrum is approximated by a sum of Gaussian functions
\begin{equation}
\phi_{\sigma}(t) = \frac{1}{N}\sum_{i=1}^{N}f(\lambda_i,t,\sigma^2)
\end{equation}
where 
\begin{equation}
f(\lambda_i,t,\sigma^2) = \frac{1}{\sigma \sqrt{2\pi}}\exp(-\frac{(t-\lambda_i)^2}{2\sigma^2})
\end{equation}
We use the Lanczos algorithm with full reorthogonalization in order to compute
eigenvalues and eigenvectors of the Hessian and to ensure orthogonality between
the different eigenvectors.  Since the Hessian is symmetric we can diagonalize
it and all eigenvalues are real.
The Lanczos algorithm is used together with the Hessian vector product for a certain number of iterations. Afterwards it returns a tridiagonal matrix $T$. This matrix is diagonalized as
\begin{equation}
T = ULU^T
\end{equation}
where $L$ is a diagonal matrix.

By setting $\omega_i = U_{1,i}^2$ and $l_i = L_{ii}$ for $i=1,2,...,m$, the resulting eigenvalues and eigenvectors are used to estimate the true eigenvalue density spectrum
\begin{equation}
\hat{\phi}^{(v_i)}(t)=\sum_{i=1}^{m}\omega_if(l_i,t,\sigma^2)
\end{equation}

\begin{equation}
\hat{\phi}_{\sigma}(t) = \frac{1}{k}\sum_{i=1}^{k}\hat{\phi}^{(v_i)}(t) 
\end{equation}
For our experiments we use the toolbox from \cite{chatzimichailidis2019gradvis} which implements the Stochastic Lanczos quadrature algorithm. This allows to inspect and visualize the spectral information from our models.

\subsection{Nudged-Adam Optimizer}
\noindent To prevent our neural network from reaching sharp minima during optimization, we remove the gradient information in the direction of high eigenvalues. This forces our network to ignore the sharpest minima entirely and instead converge into wider ones.
Inspired by \cite{jastrzebski2018relation}, we construct an optimizer based on Adam \cite{kingma2014adam} which ignores the gradient in the direction of the top-k eigenvectors. In order to achieve this, we use the existing Adam optimizer and remove the directions of steepest descent from its gradient. This means that given the top-k eigenvectors $v_i$ and the gradient $g$ we remove the eigenvector directions by 
\begin{equation}
g^* = g-\sum_{i=1}^{k}<g,v_i>v_i
\end{equation}
The resulting gradient $g^*$ is now used by the regular Adam optimizer. Using this technique one can modify a lot of different optimizers into their nudged counterpart by using $g^*$ instead of the true gradient. 

The eigenvectors are computed by using the Lanczos method together with the R-operator. This allows fast computation of eigenvalues and eigenvectors without having to store the full Hessian.

\section{\uppercase{Experiments}}

\noindent In this section, we present a set of experiments to study the loss properties and the instability issues that might occur when training a GAN. We first use the visualization toolbox to inspect the spectrum of GANs during training, and to corroborate the problematic search of an LNE. Then, we examine the k-top eigenvalues from the $G$ and $D$, and their evolution throughout the training. Finally, we introduce a novel optimizer, called nudged-Adam, to prevent mode collapse and we test its performances on several datasets to guarantee reliability across different scenarios.

\subsection{Visualizing Loss Landscape}

\noindent We start our experimental section with a loss landscape visualization that will serve us as a reference point. We believe that building a solid background will help to provide a better understanding of the non-convergent nature of GANs, in particular concerning the $G$. In order to do this, we track the spectral density throughout the entire optimization process. Our main goal here is to gather evidences from the curvature that visualizes the general problem of finding LNEs. In order to carry out these experiments, we independently analyse the loss landscapes of the $G$ and of the $D$ using their highest eigenvalue, respectively.

\begin{figure}[!h]
\centering
   \includegraphics[width=\linewidth]{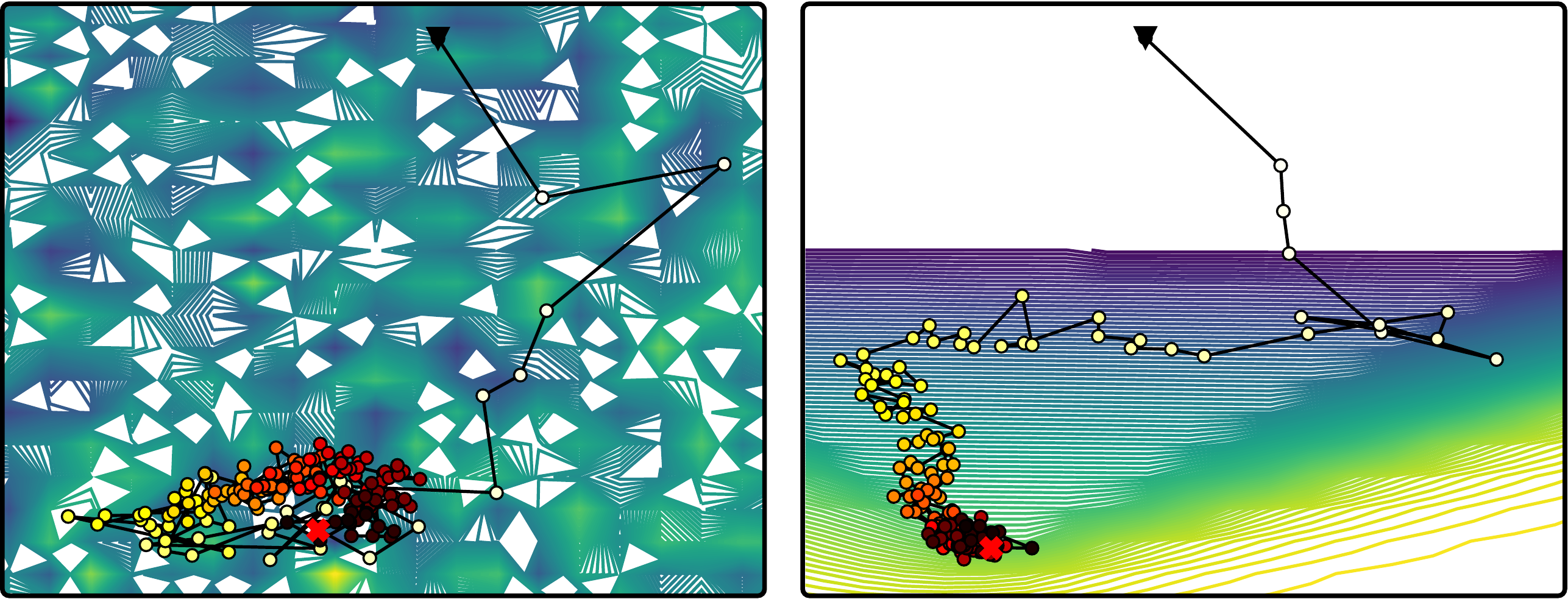}
\caption{Logarithmic loss landscapes with trajectory of the same training run, visualized along eigenvectors corresponding to the two highest eigenvalues of NSGAN on MNIST. (Left) $G$ loss landscape. (Right) $D$ loss landscape. }
\label{fig:land}
\end{figure}

We employ the toolbox from \cite{chatzimichailidis2019gradvis} to visualize the loss landscape and the trajectory of our GAN during training. In this way, we can gain some insights into the optimization process that happens underneath. Note, that to obtain the trajectory, we  project all the points of training into the 2D plane of the last epoch. Figure \ref{fig:land} shows the landscape after training for 180 epochs on the NSGAN setup. By inspecting the landscapes, we observe (1) the $D$ clearly finds a local minimum and descends towards it, and (2) the $G$ ends up in an unstable saddle point, as suggested by the irregular landscapes surrounding it. This findings agree with the aforementioned second-order literature \cite{berard2019closer}.

\subsection{Eigenvalue Analysis}

\noindent After having gained intuition of the training conditions of GANs and their problem to find and remain at an LNE, we now focus our attention on the issue of mode collapse. In particular, we provide empirical evidences of a plausible relationship between mode collapse and the behaviour of the eigenvalues. To this end, we evaluate the spectrum of our model throughout the optimization process. More specifically, we track the largest eigenvalues from the $G$ and from the $D$ for each epoch, together with the IS. 

\begin{figure*}[!t]
\begin{subfigure}{.33\linewidth}
\centering
\resizebox{\columnwidth}{!}{
	\begin{tikzpicture}
	\begin{semilogyaxis}[
	x label style={at={(axis description cs:0.5,0)}},
	xlabel=Epochs,
	legend pos=north west,
	every axis plot/.append style={semithick}]
	
		\addplot [blue, mark=] table {img/gen_kuzushiji.dat};
		\addplot [orange, mark=] table {img/disc_kuzushiji.dat};
		\addplot [black!30!green, mark=] table {img/is_kuzushiji.dat};
		\legend{gen,disc,IS}
	\end{semilogyaxis}%
	\end{tikzpicture}
	}
	\caption{Kuzushiji dataset. Correlation 0.80.}
\end{subfigure}
\begin{subfigure}{.33\linewidth}
\centering
\resizebox{\columnwidth}{!}{
	\begin{tikzpicture}
	\begin{semilogyaxis}[
	x label style={at={(axis description cs:0.5,0)}},
	xlabel=Epochs,
	legend pos=north west,
	every axis plot/.append style={semithick}]
	
		\addplot [blue, mark=] table {img/gen_fashion.dat};
		\addplot [orange, mark=] table {img/disc_fashion.dat};
		\addplot [black!30!green, mark=] table {img/is_fashion.dat};
		\legend{gen,disc,IS}
	\end{semilogyaxis}%
	\end{tikzpicture}
	}
   \caption{Fashion dataset. Correlation 0.90.}
\end{subfigure}
\begin{subfigure}{.33\linewidth}
\centering
\resizebox{\columnwidth}{!}{
	\begin{tikzpicture}
	\begin{semilogyaxis}[
	x label style={at={(axis description cs:0.5,0)}},
	xlabel=Epochs,
	legend pos=north west,
	every axis plot/.append style={semithick}]
	
		\addplot [blue, mark=] table {img/gen_emnist.dat};
		\addplot [orange, mark=] table {img/disc_emnist.dat};
		\addplot [black!30!green, mark=] table {img/is_emnist.dat};
		\legend{gen,disc,IS}
	\end{semilogyaxis}%
	\end{tikzpicture}
	}
   \caption{EMNIST dataset. Correlation 0.72.}
\end{subfigure}\\[1ex]
\caption{Evolution of the top k-eigenvalues of the Hessian from generator (gen) and discriminator (disc), and the correspondence IS over the whole training phase. The correlation score is measured between the $G$ and the $D$.}
\label{fig:gans}
\end{figure*}
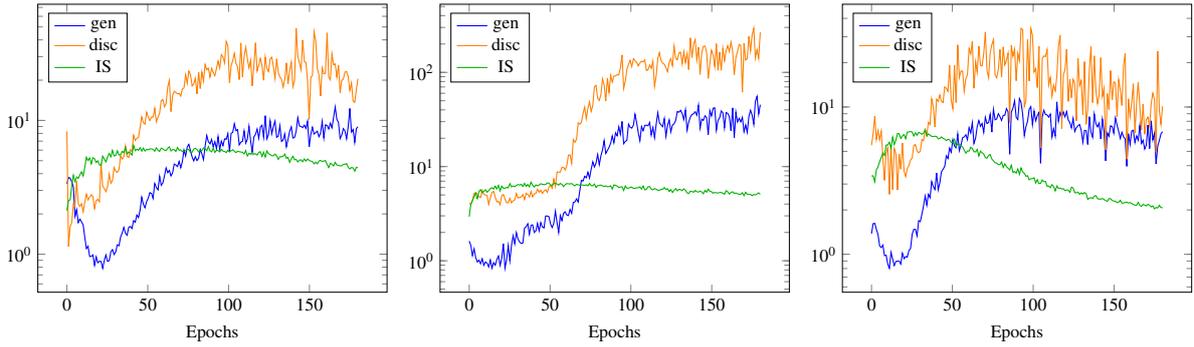

We start training and evaluating the original non-saturating GAN architecture on the MNIST, Kuzushiji, Fashion and EMNIST datasets (see Figure \ref{fig:gans}). This results in a number of patterns that are present in all experiments.
(1) The evolution of the eigenvalues of the $G$ and $D$ behave visually very similar. In particular, when $D$ exhibits an increasing tendency in its eigenvalues, the $G$ does so as well. (2) Apart from the $G$ shape of the dynamics, it is important to evaluate the local behaviour, i.e. the correlation between the $G$ and $D$. Thereby, we observe a strong correlation in all our setups ranging from 0.72 to 0.90. (3) Furthermore, there seems to exist a connection between the IS and the behaviour of the eigenvalues. When the eigenvalues have a decreasing tendency, the IS score tends to increase, while when the eigenvalues increase, the IS scores deteriorates.  Moreover, we see how all our models start to suffer from mode collapse after 25 epochs (approximately when the eigenvalues tendency changes and starts to increase). 

The empirical observations found in this analysis lead to the conclusion that eigenvalues can give an indication of the state of convergence of a GAN, as pointed out in \cite{berard2019closer}. Furthermore, we found that the eigenvalue evolution is correlated with the likely occurrence of a mode collapse event.

\begin{figure*}[!t]
\begin{subfigure}[t]{\linewidth}
\centering
   \includegraphics[width=1\linewidth]{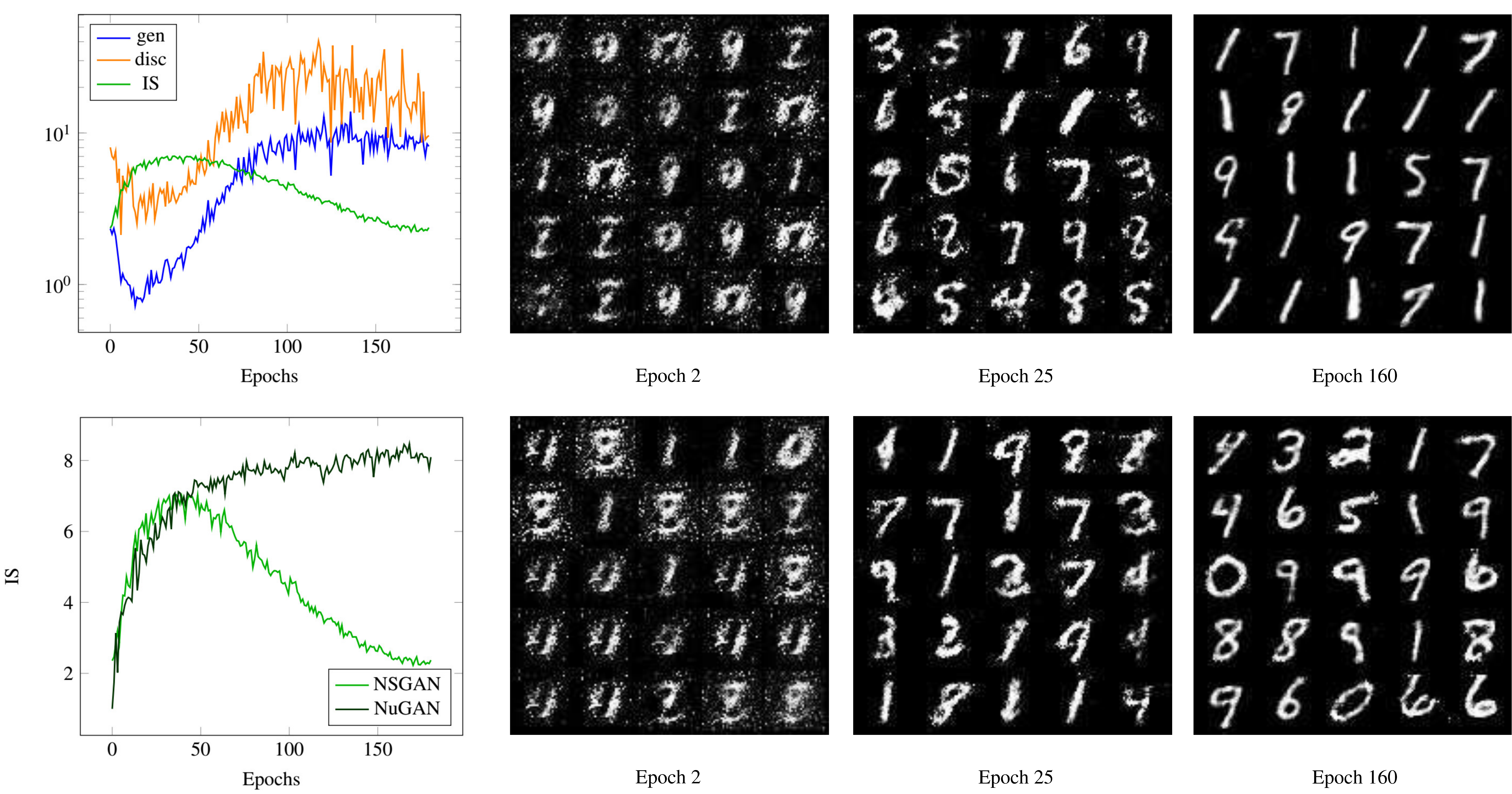}
   \label{fig:evol1}
\end{subfigure}\\[-3ex]
\caption{(First row) Evolution of the top k-eigenvalues of the Hessian, the IS and random generated samples at different epochs of NSGAN on MNIST. (Second row) Comparison of IS evolution of NSGAN and NuGAN, and random generated samples at different epochs of NuGAN.}
\label{fig:evol}
\end{figure*}

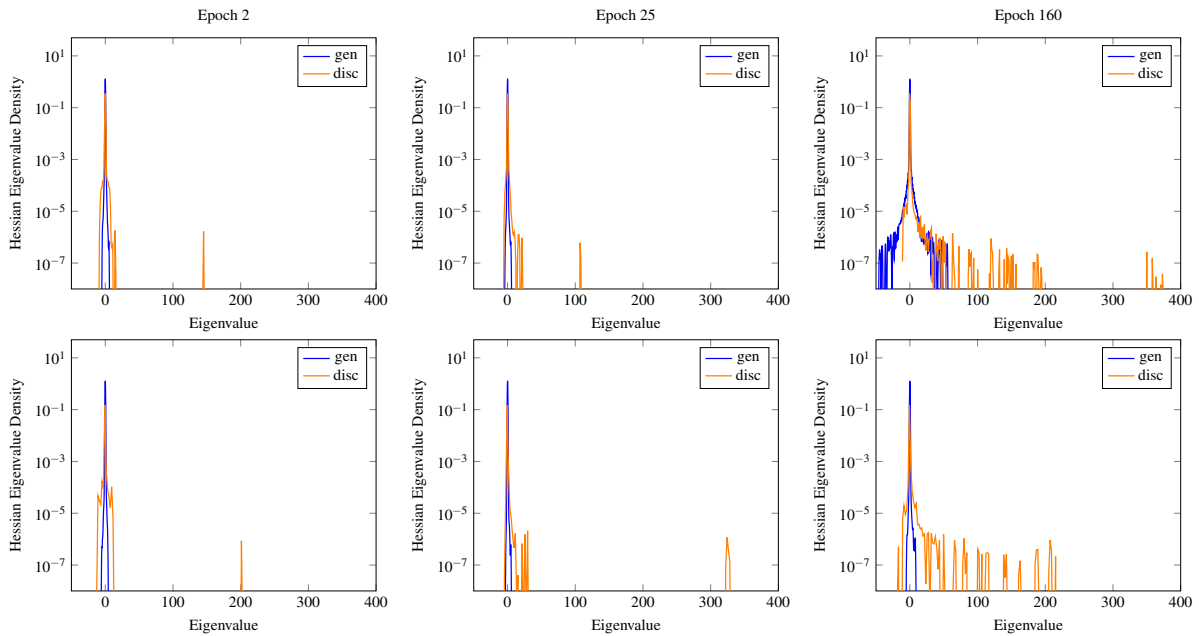
\begin{figure*}[!t]
\begin{subfigure}{.33\linewidth}
\centering
\resizebox{\columnwidth}{!}{
\begin{tikzpicture}
	\begin{semilogyaxis}[
	x label style={at={(axis description cs:0.5,0)}},
	title=Epoch 2,
	xlabel=Eigenvalue, ylabel=Hessian Eigenvalue Density,
	legend pos=north east,
	every axis plot/.append style={thick},
	xmin=-50,xmax=400,ymin=0.00000001,ymax=50]
	
	\addplot [blue, mark=] table {img/gen_spec_2.dat};
	\addplot [orange, mark=] table {img/disc_spec_2.dat};
	\legend{gen,disc}
	\end{semilogyaxis}%
\end{tikzpicture}
	}
\end{subfigure}
\begin{subfigure}{.33\linewidth}
\centering
\resizebox{\columnwidth}{!}{
\begin{tikzpicture}
	\begin{semilogyaxis}[
	x label style={at={(axis description cs:0.5,0)}},
	title=Epoch 25,
	xlabel=Eigenvalue, ylabel=Hessian Eigenvalue Density,
	legend pos=north east,
	every axis plot/.append style={thick},
	xmin=-50,xmax=400,ymin=0.00000001,ymax=50]
	
	\addplot [blue, mark=] table {img/gen_spec_25.dat};
	\addplot [orange, mark=] table {img/disc_spec_25.dat};
	\legend{gen,disc}
	\end{semilogyaxis}%
\end{tikzpicture}
	}
\end{subfigure}
\begin{subfigure}{.33\linewidth}
\centering
\resizebox{\columnwidth}{!}{
\begin{tikzpicture}
	\begin{semilogyaxis}[
	x label style={at={(axis description cs:0.5,0)}},
	title=Epoch 160,
	xlabel=Eigenvalue, ylabel=Hessian Eigenvalue Density,
	legend pos=north east,
	every axis plot/.append style={thick},
	xmin=-50,xmax=400,ymin=0.00000001,ymax=50]
	
	\addplot [blue, mark=] table {img/gen_spec_160.dat};
	\addplot [orange, mark=] table {img/disc_spec_160.dat};
	\legend{gen,disc}
	\end{semilogyaxis}%
\end{tikzpicture}
	}
\end{subfigure}

\begin{subfigure}{.33\linewidth}
\centering
\resizebox{\columnwidth}{!}{
\begin{tikzpicture}
	\begin{semilogyaxis}[
	x label style={at={(axis description cs:0.5,0)}},
	xlabel=Eigenvalue, ylabel=Hessian Eigenvalue Density,
	legend pos=north east,
	every axis plot/.append style={thick},
	xmin=-50,xmax=400,ymin=0.00000001,ymax=50]
	\addplot [blue, mark=] table {img/gen_spec_adam_2.dat};
	\addplot [orange, mark=] table {img/disc_spec_adam_2.dat};
	\legend{gen,disc}
	\end{semilogyaxis}%
\end{tikzpicture}
	}
\end{subfigure}
\begin{subfigure}{.33\linewidth}
\centering
\resizebox{\columnwidth}{!}{
\begin{tikzpicture}
	\begin{semilogyaxis}[
	x label style={at={(axis description cs:0.5,0)}},
	xlabel=Eigenvalue, ylabel=Hessian Eigenvalue Density,
	legend pos=north east,
	every axis plot/.append style={thick},
	xmin=-50,xmax=400,ymin=0.00000001,ymax=50]
	\addplot [blue, mark=] table {img/gen_spec_adam_25.dat};
	\addplot [orange, mark=] table {img/disc_spec_adam_25.dat};
	\legend{gen,disc}
	\end{semilogyaxis}%
\end{tikzpicture}
	}
\end{subfigure}
\begin{subfigure}{.33\linewidth}
\centering
\resizebox{\columnwidth}{!}{
\begin{tikzpicture}
	\begin{semilogyaxis}[
	x label style={at={(axis description cs:0.5,0)}},
	xlabel=Eigenvalue, ylabel=Hessian Eigenvalue Density,
	legend pos=north east,
	every axis plot/.append style={thick},
	xmin=-50,xmax=400,ymin=0.00000001,ymax=50]
	
	\addplot [blue, mark=] table {img/gen_spec_adam_160.dat};
	\addplot [orange, mark=] table {img/disc_spec_adam_160.dat};
	\legend{gen,disc}
	\end{semilogyaxis}%
\end{tikzpicture}
	}
\end{subfigure}
\caption{Plots of the whole spectrum of the Hessian at different stage of the training on MNIST. (First row) Results on NSGAN: we can identify an abnormal behaviour (mode collapse) in the generator at epoch 160. (Second row) Results on NuGAN: the spectrum remains stable during the whole training. We can observe how the $D$ for both cases finds local minima, while the $G$ remains all the time in a saddle point.}
\label{fig:spec}
\end{figure*}

\subsection{Combating Mode Collapse}

\noindent In the last section we have seen that the growth of Hessian eigenvalues during the training of a GAN correlates to the occurrence of mode collapse. In order to remove this undesirable effect, we train our NSGAN with a nudged-Adam optimizer (referred to as NuGAN), which is inspired by \cite{jastrzebski2018relation}.
Figure~\ref{fig:evol} shows the results together with some visual samples generated at different training epochs. We observe that NuGAN achieves a much more stable IS, and this is also displayed on the generated samples. While NSGAN suffers from mode collapse, NuGAN does not (see samples on epoch 160). This shows a clear relationship between the behaviour of the IS score and the occurrence of mode collapse.

Figure~\ref{fig:spec} shows the full spectrum of the Hessian at different stages of the training. A remarkable observation here is the present of negative eigenvalues for the $G$ for both optimizers. This indicates that the critical point reached during training is not an LNE (c.f. Formula~\ref{eqn:LNE}). Rather, the $G$ reaches only a saddle point in all cases. On the other hand, the $D$ seems to converge to a sharp local minimum when using plain GDA. In fact, it seems that the longer training lasts the sharper the minimum gets. The $D$ of our NuGAN however, reaches a much flatter minimum, which can be seen by the presents of much smaller eigenvalues towards the end of training.

A second interesting observation is the connection between the spectrum of the $G$ and the mode collapse. In particular, we observe the occurrence of mode collapse when the spectrum spreads significantly (see first row from Figure~\ref{fig:spec}). On the other hand, the spectral evolution of our NuGAN (see second row from Figure~\ref{fig:spec}) does not display any anomaly for the $G$, and indeed, no mode collapse event occurrences. 
In Table~\ref{table:is} we show more quantitative results supporting the benefit of our nudged-Adam optimizer approach. There we show the IS for both optimizers evaluated of 4 different datasets. Notice that in all cases, our method achieves a higher mean and maximum score than the NSGAN baseline. These quantitative results together with the visual inspection of the image quality suggest that our NuGAN algorithm has a direct influence on the behavior of the eigenvalues and the loss landscape of our adversarial model, resulting in the avoidance of mode collapse.

\begin{table}[!h]
\caption{Mean and max IS from the different datasets and methods (with and without mode collapse). Higher values are better.}
\centering
\begin{tabular}{ccccc}
\hline
Methods  & \multicolumn{2}{c}{NSGAN} & \multicolumn{2}{c}{NuGAN} \\
\hline
   & mean & max & mean & max \\
\hline
MNIST & 4.30 & 7.03 & 7.14 & 8.46\\
Kuzushiji & 5.24 & 6.50 & 6.12 & 7.20\\
Fashion & 5.74 & 6.82 & 6.35 & 7.20\\
EMNIST & 3.77 & 7.02 & 8.53 & 7.67\\
\hline
\end{tabular}
\label{table:is}
\end{table}

Overall, we can summarize that the algorithm does not converge to an LNE, while still achieving good results w.r.t. the evaluation metric (IS). This raises the question whether convergences to an LNE is actually needed in order to achieve good generator performance of a GAN.

\section{\uppercase{Conclusions}}

\noindent In this work, we investigate instabilities that occur during the training of GANs, focusing particularly on the issue of mode collapse. To do this, we analyse the loss surfaces of the $G$ and $D$ neural networks, using second-order gradient information, with special attention on the Hessian eigenvalues. Hereby, we empirically show that there exists a correlation between the stability of training and the eigenvalues of the generator network. In particular, we observe that large eigenvalues, which may be an indication of the convergence towards a sharp minimum, correlate well with the occurrence of mode collapse. Motivated by this observation, we introduce a novel optimization algorithm that uses second-order information to steer away from sharp optima, thereby preventing the occurrence of mode collapse. Our findings suggest that the investigation of generalization properties of GANs, e.g. by analysing the flatness of the optima found during training, is a promising approach to progress towards more stable GAN training as well.

\bibliographystyle{apalike}
{\small
\bibliography{example}}

\begin{thebibliography}{}

\bibitem[Abdal et~al., 2019]{abdal2019image2stylegan}
Abdal, R., Qin, Y., and Wonka, P. (2019).
\newblock Image2stylegan: How to embed images into the stylegan latent space?
\newblock In {\em Proceedings of the IEEE international conference on computer
  vision}, pages 4432--4441.

\bibitem[Adolphs et~al., 2018]{adolphs2018local}
Adolphs, L., Daneshmand, H., Lucchi, A., and Hofmann, T. (2018).
\newblock Local saddle point optimization: A curvature exploitation approach.
\newblock {\em arXiv preprint arXiv:1805.05751}.

\bibitem[Alain et~al., 2019]{alain2019negative}
Alain, G., Roux, N.~L., and Manzagol, P.-A. (2019).
\newblock Negative eigenvalues of the hessian in deep neural networks.
\newblock {\em arXiv preprint arXiv:1902.02366}.

\bibitem[Arjovsky et~al., 2017]{arjovsky2017wasserstein}
Arjovsky, M., Chintala, S., and Bottou, L. (2017).
\newblock Wasserstein gan.
\newblock {\em arXiv preprint arXiv:1701.07875}.

\bibitem[Berard et~al., 2019]{berard2019closer}
Berard, H., Gidel, G., Almahairi, A., Vincent, P., and Lacoste-Julien, S.
  (2019).
\newblock A closer look at the optimization landscapes of generative
  adversarial networks.
\newblock {\em arXiv preprint arXiv:1906.04848}.

\bibitem[Chatzimichailidis et~al., 2019]{chatzimichailidis2019gradvis}
Chatzimichailidis, A., Keuper, J., Pfreundt, F.-J., and Gauger, N.~R. (2019).
\newblock Gradvis: Visualization and second order analysis of optimization
  surfaces during the training of deep neural networks.
\newblock In {\em 2019 IEEE/ACM Workshop on Machine Learning in High
  Performance Computing Environments (MLHPC)}, pages 66--74. IEEE.

\bibitem[Chaudhari et~al., 2019]{chaudhari2019entropy}
Chaudhari, P., Choromanska, A., Soatto, S., LeCun, Y., Baldassi, C., Borgs, C.,
  Chayes, J., Sagun, L., and Zecchina, R. (2019).
\newblock Entropy-sgd: Biasing gradient descent into wide valleys.
\newblock {\em Journal of Statistical Mechanics: Theory and Experiment},
  2019(12):124018.

\bibitem[Dinh et~al., 2017]{LaurentSharp}
Dinh, L., Pascanu, R., Bengio, S., and Bengio, Y. (2017).
\newblock Sharp minima can generalize for deep nets.
\newblock {\em CoRR}, abs/1703.04933.

\bibitem[Draxler et~al., 2018]{draxler2018essentially}
Draxler, F., Veschgini, K., Salmhofer, M., and Hamprecht, F.~A. (2018).
\newblock Essentially no barriers in neural network energy landscape.
\newblock {\em arXiv preprint arXiv:1803.00885}.

\bibitem[Durall et~al., 2020]{durall2020watch}
Durall, R., Keuper, M., and Keuper, J. (2020).
\newblock Watch your up-convolution: Cnn based generative deep neural networks
  are failing to reproduce spectral distributions.
\newblock In {\em Proceedings of the IEEE/CVF Conference on Computer Vision and
  Pattern Recognition}, pages 7890--7899.

\bibitem[Durall et~al., 2019]{durall2019object}
Durall, R., Pfreundt, F.-J., K{\"o}the, U., and Keuper, J. (2019).
\newblock Object segmentation using pixel-wise adversarial loss.
\newblock In {\em German Conference on Pattern Recognition}, pages 303--316.
  Springer.

\bibitem[Fiez et~al., 2019]{fiez2019convergence}
Fiez, T., Chasnov, B., and Ratliff, L.~J. (2019).
\newblock Convergence of learning dynamics in stackelberg games.
\newblock {\em arXiv preprint arXiv:1906.01217}.

\bibitem[Goodfellow et~al., 2014a]{goodfellow2014generative}
Goodfellow, I., Pouget-Abadie, J., Mirza, M., Xu, B., Warde-Farley, D., Ozair,
  S., Courville, A., and Bengio, Y. (2014a).
\newblock Generative adversarial nets.
\newblock In {\em Advances in neural information processing systems}, pages
  2672--2680.

\bibitem[Goodfellow et~al., 2014b]{goodfellow2014qualitatively}
Goodfellow, I.~J., Vinyals, O., and Saxe, A.~M. (2014b).
\newblock Qualitatively characterizing neural network optimization problems.
\newblock {\em arXiv preprint arXiv:1412.6544}.

\bibitem[Gulrajani et~al., 2017]{gulrajani2017improved}
Gulrajani, I., Ahmed, F., Arjovsky, M., Dumoulin, V., and Courville, A.~C.
  (2017).
\newblock Improved training of wasserstein gans.
\newblock In {\em Advances in neural information processing systems}, pages
  5767--5777.

\bibitem[Hochreiter and Schmidhuber, 1997]{hochreiter1997flat}
Hochreiter, S. and Schmidhuber, J. (1997).
\newblock Flat minima.
\newblock {\em Neural Computation}, 9(1):1--42.

\bibitem[Iizuka et~al., 2017]{iizuka2017globally}
Iizuka, S., Simo-Serra, E., and Ishikawa, H. (2017).
\newblock Globally and locally consistent image completion.
\newblock {\em ACM Transactions on Graphics (ToG)}, 36(4):107.

\bibitem[Izmailov et~al., 2018]{izmailov2018averaging}
Izmailov, P., Podoprikhin, D., Garipov, T., Vetrov, D., and Wilson, A.~G.
  (2018).
\newblock Averaging weights leads to wider optima and better generalization.
\newblock {\em arXiv preprint arXiv:1803.05407}.

\bibitem[Jastrzebski et~al., 2018]{jastrzebski2018relation}
Jastrzebski, S., Kenton, Z., Ballas, N., Fischer, A., Bengio, Y., and Storkey,
  A. (2018).
\newblock On the relation between the sharpest directions of dnn loss and the
  sgd step length.
\newblock {\em arXiv preprint arXiv:1807.05031}.

\bibitem[Karras et~al., 2020]{karras2020analyzing}
Karras, T., Laine, S., Aittala, M., Hellsten, J., Lehtinen, J., and Aila, T.
  (2020).
\newblock Analyzing and improving the image quality of stylegan.
\newblock In {\em Proceedings of the IEEE/CVF Conference on Computer Vision and
  Pattern Recognition}, pages 8110--8119.

\bibitem[Keskar et~al., 2016]{keskar2016large}
Keskar, N.~S., Mudigere, D., Nocedal, J., Smelyanskiy, M., and Tang, P. T.~P.
  (2016).
\newblock On large-batch training for deep learning: Generalization gap and
  sharp minima.
\newblock {\em arXiv preprint arXiv:1609.04836}.

\bibitem[Kingma and Ba, 2014]{kingma2014adam}
Kingma, D.~P. and Ba, J. (2014).
\newblock Adam: A method for stochastic optimization.
\newblock {\em arXiv preprint arXiv:1412.6980}.

\bibitem[Lanczos, 1950]{lanczos1950iteration}
Lanczos, C. (1950).
\newblock {\em An iteration method for the solution of the eigenvalue problem
  of linear differential and integral operators}.
\newblock United States Governm. Press Office Los Angeles, CA.

\bibitem[Mescheder et~al., 2017]{mescheder2017numerics}
Mescheder, L., Nowozin, S., and Geiger, A. (2017).
\newblock The numerics of gans.
\newblock In {\em Advances in Neural Information Processing Systems}, pages
  1825--1835.

\bibitem[Nagarajan and Kolter, 2017]{nagarajan2017gradient}
Nagarajan, V. and Kolter, J.~Z. (2017).
\newblock Gradient descent gan optimization is locally stable.
\newblock In {\em Advances in neural information processing systems}, pages
  5585--5595.

\bibitem[Radford et~al., 2015]{radford2015unsupervised}
Radford, A., Metz, L., and Chintala, S. (2015).
\newblock Unsupervised representation learning with deep convolutional
  generative adversarial networks.
\newblock {\em arXiv preprint arXiv:1511.06434}.

\bibitem[Sagun et~al., 2016]{sagun2016eigenvalues}
Sagun, L., Bottou, L., and LeCun, Y. (2016).
\newblock Eigenvalues of the hessian in deep learning: Singularity and beyond.
\newblock {\em arXiv preprint arXiv:1611.07476}.

\bibitem[Salimans et~al., 2016]{salimans2016improved}
Salimans, T., Goodfellow, I., Zaremba, W., Cheung, V., Radford, A., and Chen,
  X. (2016).
\newblock Improved techniques for training gans.
\newblock In {\em Advances in neural information processing systems}, pages
  2234--2242.

\bibitem[Xue et~al., 2018]{xue2018segan}
Xue, Y., Xu, T., Zhang, H., Long, L.~R., and Huang, X. (2018).
\newblock Segan: Adversarial network with multi-scale l 1 loss for medical
  image segmentation.
\newblock {\em Neuroinformatics}, 16(3-4):383--392.

\bibitem[Yu et~al., 2019]{yu2019free}
Yu, J., Lin, Z., Yang, J., Shen, X., Lu, X., and Huang, T.~S. (2019).
\newblock Free-form image inpainting with gated convolution.
\newblock In {\em Proceedings of the IEEE International Conference on Computer
  Vision}, pages 4471--4480.

\end{thebibliography}

\end{document}